\documentclass[10pt]{article}
\usepackage{makeidx}
\usepackage{multirow}
\usepackage{multicol}
\usepackage[dvipsnames,svgnames,table]{xcolor}
\usepackage{graphicx}
\usepackage{epstopdf}
\usepackage{ulem}
\usepackage{hyperref}
\usepackage{amsmath}
\usepackage{amssymb}
\usepackage[multi-part-units=single]{siunitx}
\author{DrX}
\title{Towards Head Motion Compensation Using Multi-Scale Convolutional Neural Networks}
\usepackage[paperwidth=595pt,paperheight=842pt,top=70pt,right=56pt,bottom=70pt,left=56pt]{geometry}

\makeatletter
	{\par\setlength{\parindent}{#3}
	\setlength{\leftmargin}{#1}       \setlength{\rightmargin}{#2}%
	\advance\linewidth -\leftmargin       \advance\linewidth -\rightmargin%
	\advance\@totalleftmargin\leftmargin  \@setpar{{\@@par}}%
	\parshape 1\@totalleftmargin \linewidth\ignorespaces}{\par}%
\makeatother 

\begin{document}

\pagestyle{empty}

\begin{center}
\textbf{\Large{Towards Head Motion Compensation Using Multi-Scale Convolutional Neural Networks}}
\end{center}

\begin{center}
\textit{O. Rajput${^1}$${^*}$, N. Gessert${^1}$${^*}$, M. Gromniak${^1}$, L. Matth\"aus${^2}$, A. Schlaefer${^1}$}
\end{center}

\begin{center}
\textit{${^1}$ Institute of Medical Technology, Hamburg University of Technology, Hamburg, Germany}\\
\textit{${^2}$ eemagine Medical Imaging Solutions GmbH, Berlin, Germany}\\
\textit{${^*}$ Both authors contributed equally.}\\
\end{center}

\begin{center}
\text{Contact: omer.rajput@tuhh.de, nils.gessert@tuhh.de}
\end{center}

{\raggedright
\textbf{\textit{Abstract}}
}
\\
\\
\textit{Head pose estimation and tracking is useful in variety of medical applications. With the advent of RGBD cameras like Kinect, it has become feasible to do markerless tracking by estimating the head pose directly from the point clouds. One specific medical application is robot assisted transcranial magnetic stimulation (TMS) where any patient motion is compensated with the help of a robot. For increased patient comfort, it is important to track the head without markers. In this regard, we address the head pose estimation problem using two different approaches. In the first approach, we build upon the more traditional approach of model based head tracking, where a head model is morphed according to the particular head to be tracked and the morphed model is used to track the head in the point cloud streams.
In the second approach, we propose a new multi-scale convolutional neural network architecture for more accurate pose regression. Additionally, we outline a systematic data set acquisition strategy using a head phantom mounted on the robot and ground-truth labels generated using a highly accurate tracking system.
}

\section{\hspace{14pt}Problem}

Motion compensation is necessary in various medical procedures including transcranial magnetic stimulation (TMS). TMS is a method used to stimulate cortical regions of the brain, e.g., to treat tinnitus. The necessity of motion compensation arises from the fact that patients are typically not static due to, e.g., breathing and other involuntary movements. Without the motion compensation, the patient’s mobility has to be restricted with unpleasant aides like plaster casts \cite{weltens1995comparison}. 
For motion compensation in TMS, the patient is tracked with a tracking system and the robot compensates the head motion. Previously, TMS with motion compensating robots have been proposed using marker-based tracking which involves placing/sticking retro-reflective markers on a patient's head \cite{Richter.2013}. The marker placement typically causes irritation to the treated patients and their placement is likely to be affected when the patients scratch or touch the markers, thus compromising the overall accuracy of the compensation.

Therefore, markerless tracking approaches based on adjusted models have been proposed where a subject-specific template is tracked with an Iterative Closest Point (ICP) algorithm \cite{breitenstein2008real}. The model-based approach splits the problem into two major subproblems: An offline fitting step to create a subject-specific face template \cite{paysan20093d} from noisy sensor data and its registration to the sensor frames using a variant of the Iterative Closest Point algorithm \cite{rusinkiewicz2001efficient} during the actual pose estimation phase. 
Recently, CNNs have also been proposed for head pose estimation "in the wild" \cite{patacchiola2017head} and for automotive driver head pose estimation \cite{schwarz2017driveahead}. Often, these approaches use coarse, discrete poses which is insufficient for TMS application. 

Thus, we propose CNN-based head pose estimation from RGBD+IR images for TMS-like applications using a data acquisition setup with a highly accurate tracking camera for ground-truth annotation. A robot moves a head phantom for automatic data acquisition which allows for generation of large datasets to study head pose estimation approaches. Based on the acquired datasets, we propose a multi-scale CNN architecture for accurate pose regression. We compare this method to a single-path CNN and a classical head model-based approach.

\section{\hspace{14pt}Materials and Methods}

\subsection{\hspace{12pt}Head Phantom Dataset Generation}
\begin{figure}
	\includegraphics[width=\textwidth]{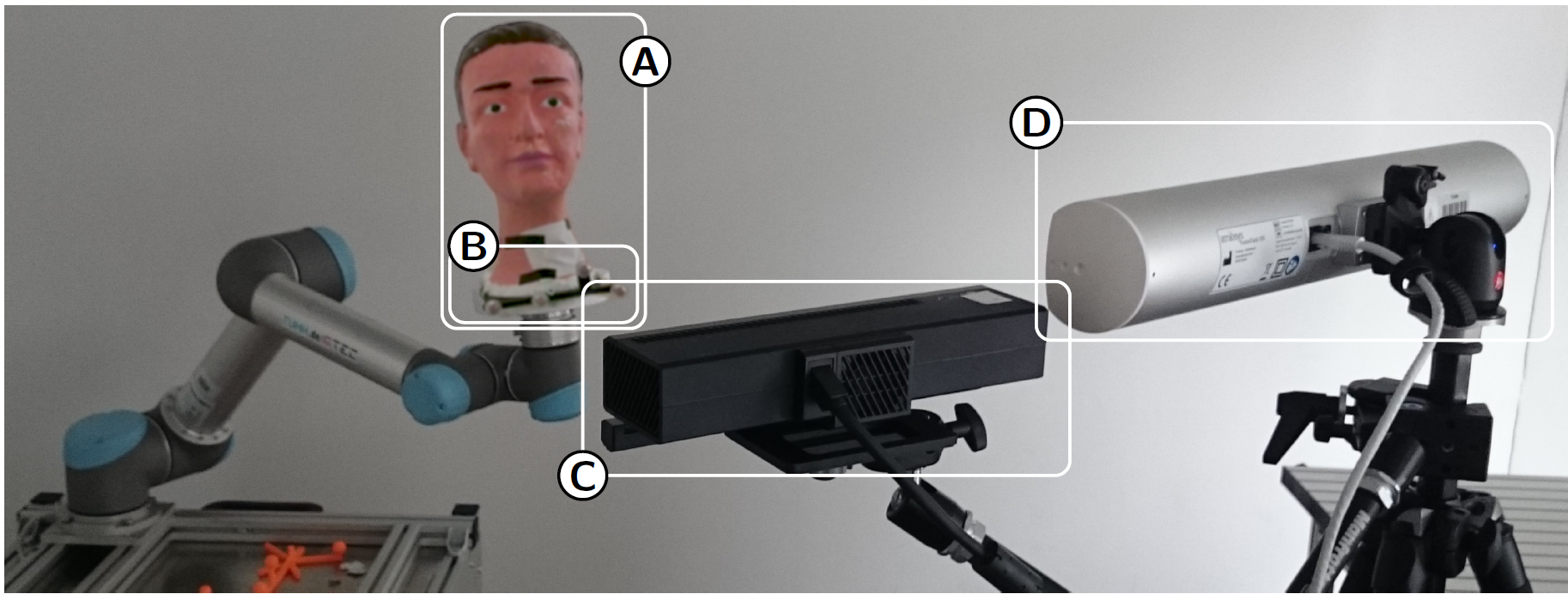}
	\caption{Acquisition setup. Head phantom (A), head marker geometry (B), Microsoft Kinect v2 (C), optical tracking system (D)}
	\label{setup}
\end{figure}

A dataset of the head phantom is generated to systematically cover predefined distances and orientations towards the Kinect sensor. Universal Robot's UR5 robot is used to systematically move the head in the Kinect's FoV. In addition to Kinect, a highly accurate optical tracking system `Atracsys fusionTrack 500' is used to track the head as well to record the corresponding ground truth head poses. The tracking system tracks a retroreflective markers based rigid geometries, therefore, one of such marker geometries is fixed to the phantom’s neck. This was done such that the markers do not interfere/occlude the head/face of the phantom as shown in Figure \ref{setup}. The robot's trajectory is planned such that the overlap between robot workspace and Kinect FoV is discretized into a grid. At each position in the grid, the robot rotates the head phantom covering combinations of rotations in a range from +/-20 degrees for yaw, pitch and roll axes, respectively. Also, the order of rotations are randomized at each grid position. It is important to note that the dataset is recorded continuously, therefore, even for the coarse planned grid we have a large dataset covering a variety of head orientations in different parts of the FoV. 
Randomized robot movement leads to a dataset of approximately $\num{45000}$ images. We use $\SI{80}{\percent}$ for training/validation and $\SI{20}{\percent}$ for testing.


\subsection{\hspace{12pt}CNN Architecture}

\begin{figure}
	\includegraphics[width=\textwidth]{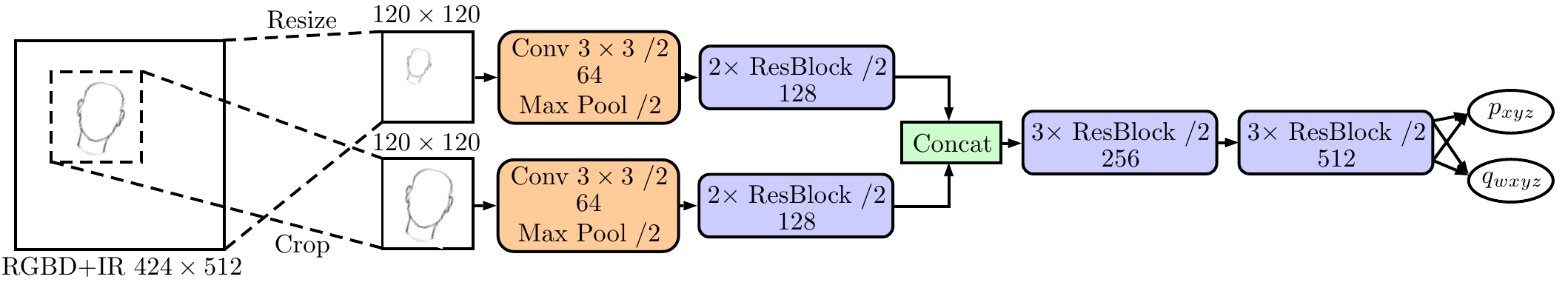}
    \caption{The proposed multi-scale CNN architecture. One path receives a downsampled version of the entire image, the other path receives a cropped version, containing mostly the head itself. Cropping is performed through a face detecting classifier. \textit{ResBlock} refers to residual blocks with bottlenecks, following \cite{He.2016}. In each block, the number of output feature maps is denoted. $/2$ denotes spatial downsampling with a convolutional stride of $2$. $p_{xyz}$ denotes the position and $q_{wxyz}$ denotes the orientation, expressed as a quaternion.}
    \label{fig:cnn}
\end{figure}

Our proposed CNN architecture is shown in Figure~\ref{fig:cnn}. The architecture follows a multi-scale approach with two paths. One path receives the entire, downsampled Kinect image. The second path receives a cropped image with a smaller FoV but higher resolution as its input. The cropped image is obtained by using a face detecting classifier \cite{deniz2011face}. In this way, the CNN is provided both global position information and fine-grained high resolution information for more accurate pose estimation. Both paths are fused inside the network through feature map concatenation. The architecture is built with ResNet-like blocks with bottlenecks for higher efficiency \cite{He.2016}.

The cropped image is obtained by using a face detecting classifier applied on the RGB image. The face detector is based on the histogram of oriented gradients (HOG) features combined with a linear classifier (like LDA). The classification uses a sliding window detection scheme on multiple scales of the image to detect faces of different sizes \cite{deniz2011face}. The classifier provides a region of interest containing the face which is used for cropping. A 120x120 section around the ROI's center is cropped from the original image to get the cropped image for the CNN.

For training, we consider all five image channels (RGB+Depth+IR) as channels at the model input. The RGB images were registered to the IR-based representations. The network performs regression by outputting seven continuous values which represent position and orientation (quaternions). Images and target poses are scaled to a range of $[0,1]$. We minimize the mean squared error between predictions and targets using the Adam algorithm \cite{Kingma.2014}. We implement our model in Tensorflow \cite{Abadi.2016}.

We compare this approach to a standard single-path CNN that only takes the entire Kinect image as its input. We do not consider a single-path CNN with the cropped images, as they do not contain global position information. Furthermore, we consider a classical head model based approach for tracking, in which the head template is expressed in terms of a statistical face model \cite{paysan20093d}. The face model is deformed using a head scan to create a head template corresponding to the specific head shape to be tracked. An optical tracking system and a pointer probe are used to sample a subject's facial surface.  The tracking algorithm is initialized by an RGB face detector and uses the Iterative Closest Point algorithm with subject-specific template for pose refinement.

Since a marker geometry is rigidly attached to the head, the head does not need to be static during the scan. The transformations from tracking camera to head marker ${}^C T_{M_i}$ and the transformations from tracking camera to pointer probe ${}^C T_{P_i}$ are synchronously recorded for $n$ samples. The translational part of the relative transformations ${}^M T_{P_i} = ({}^C T_{M_i})^{-1} {}^C T_{P_i} , i \in 1, \cdots, n$ from head marker to pointer probe can be compiled into a three-dimensional head scan.

\section{\hspace{14pt}Results}

\begin{table}
	\centering
	\begin{tabular}{l l l l}
	 & Model-Based & Single-Path & Multi-Path \\ \hline \\
	Position Error &  $\SI{4.16 \pm 2.21}{\milli\metre}$ & $\SI{2.23 \pm 1.38}{\milli\metre}$ & $\SI{1.86 \pm 1.29}{\milli\metre}$  \\
	Orientation Error & $\SI{4.63 \pm 1.42}{\degree}$ & $\SI{1.27 \pm 0.78}{\degree}$  & $\SI{0.676 \pm 0.35}{\degree}$  \\ \hline \\
	\end{tabular}
	\caption{Mean and standard deviation of the errors for the different approaches. Errors are calculated by considering $T_P^{-1}T_{GT} = I_E \stackrel{!}{=} I$ where $T_P$ is the predicted pose and $T_{GT}$ the ground-truth. The result $I_E$ should be approximately equal to the identity matrix $I$. Position error is the norm of the translational part of $I_E$ and orientation error is the angle of the rotational part of $I_E$ in axis-angle representation.}
	\label{tab:results}
\end{table}

\begin{table}
	\centering
	\begin{tabular}{l l l l}
	 & Model-Based & Single-Path & Multi-Path \\ \hline \\
	Setup Time &  $\SI{15}{\min}$ & $\SI{6}{\hour}$ & $\SI{14}{\hour}$  \\
	Processing Time & $\SI{27.1\pm 8.6}{\milli\second}$ & $\SI{5.1 \pm 0.3}{\milli\second}$  & $\SI{6.4 \pm 0.3}{\milli\second}$  \\ \hline \\
	\end{tabular}
	\caption{Setup and processing time for the three approaches. For the model-based approach, setup time refers to the time required to generate the head template. For the CNN models, setup time refers to the training time needed. Processing time denotes the time that is required for processing a single image for tracking.}
	\label{tab:results_runtime}
\end{table}

The results are shown in Table~\ref{tab:results}. The CNN-based models clearly outperform the classical model-based approach. Moreover, the two-path architecture achieves higher rotational accuracy than the single-path model. 

Additionally, we provide results for the training and runtime of each approach. The results are shown in Table~\ref{tab:results_runtime}. The training time for model based approach is required to setup the subject-specific head template as described in the previous section. The CNN requires significantly more setup time due to model training, compared to the classic approach. However, for application, the processing time for CNN inference for each image is four times faster than the model-based tracker.

\section{\hspace{14pt}Discussion}

We propose a new multi-path CNN architecture for head pose estimation, that takes both global and local context into account. Our results in Table~\ref{tab:results} show, that the method outperforms a classic model-based approach. This indicates that directly learning features from color and depth images is advantageous compared to point-cloud-based models that only rely on depth. Note, that these results are limited as they are shown for one head phantom. Although generalization in between heads does not appear to be problematic for similar problems \cite{schwarz2017driveahead}, this should be addressed in future research. Moreover, we show that the multi-scale approach outperforms a single-path model that only takes the entire image as the input. The higher resolution crop appears to provide more accurate information for orientation learning. 
The setup and processing times shown in Table~\ref{tab:results_runtime} show that the CNN models need significantly longer preparation time, as the model has to be trained on a large dataset. However, when being applied for head pose estimation, the CNNs are significantly faster than the classic approach in terms of processing time. Thus, our new CNN approach is well suited for practical applications, as fast feedback on head pose changes during a TMS scenario can allow for better motion compensation strategies.

\section{\hspace{14pt}Conclusion}

We address the problem of head pose estimation for motion compensation in TMS scenarios. For this purpose, we introduce an automatic data acquisition setup with highly accurate ground-truth annotation in order to obtain large datasets to study pose estimation methods. We introduce a new multi-scale CNN architecture for accurate pose regression. We show that our method outperforms a classic approach. Moreover, our new multi-scale CNN design improves performance over a simple single-path model. In this way, we show the successful integration of prior, coarse pose estimation methods in our new deep learning-based setup. Additionally, processing duration improves significantly, which implies that real-time estimation is feasible. For future work, this approach can be extended to multiple head phantoms and potential trials with human subjects.

\section*{Acknowledgements}

This work was partially supported by grant number ZF 4026301KW5.

\bibliographystyle{spmpsci} 
\small      
\bibliography{egbib}

\end{document}